# SUPER: A Novel Lane Detection System

Pingping Lu[1], Chen Cui[2], Shaobing Xu[1], Huei Peng[1], Fan Wang[2]

*Abstract*—AI-based lane detection algorithms were actively studied over the last few years. Many have demonstrated superior performance compared with traditional feature-based methods. The accuracy, however, is still generally in the low 80% or high 90%, or even lower when challenging images are used. In this paper, we propose a real-time lane detection system, called Scene Understanding Physics-Enhanced Real-time (SUPER) algorithm. The proposed method consists of two main modules: 1) a hierarchical semantic segmentation network as the scene feature extractor and 2) a physics enhanced multi-lane parameter optimization module for lane inference. We train the proposed system using heterogeneous data from Cityscapes, Vistas and Apollo, and evaluate the performance on four completely separate datasets (that were never seen before), including Tusimple, Caltech, URBAN KITTI-ROAD, and X-3000. The proposed approach performs the same or better than lane detection models already trained on the same dataset, and performs well even on datasets it was never trained on. Real-world vehicle tests were also conducted. Preliminary test results show promising real-time lane-detection performance compared with the Mobileye.

*Index Terms*—Lane detection, semantic segmentation, scene understanding, optimization, camera based perception

## I. INTRODUCTION

Accurate and reliable lane detection is a critical feature for Lane Keeping (LK), Lane Change Automation (LCA) and Lane Departure Warning (LDW) functions. Lane detection research can be traced back to the 1980's [1]. After the turn of the century, LDW and LK have been commercialized, some vehicles even have LCA. The self-driving challenge launched by DARPA (2004-2012) [2] and early ADAS products, e.g., the Mobileye [3], further advanced the development of lane detection systems. However, due to the diverse appearances due to adverse lighting/weather conditions and presence of other objects, lane detection is still a challenging problem. Precision and recall rate of lane markings are in the low 80% or high 90% in most papers in recent literature, still not good enough for safe and reliable driving in the real world.

Lane detection can be achieved by using monocular cameras, stereo cameras, lidars, etc.[4]. Cameras are most popular due to their rich content features and affordable cost. This paper mainly focuses on monocular cameras, but the concept could work with other sensors—albeit with significant additional work.

### A. Motivation and Literature Review

Many feature-based methods use the generic framework summarized in [4]. They decompose road/lane detection methods into several modules: image pre-processing, feature extraction, model fitting, image to world correspondence and time integration. Not all lane detection methods contain all these modules, but in general most papers contribute to one or more of these elements [5-8]. Deep Learning (DL) presents a newer data-driven approach and achieves better performance than most feature-based methods right out of the gate [9-15]. Although DL systems achieved superior performance in many applications, they are often used as a "blackbox" and their performance has no guarantee and behavior hard to explain [16]. This limits their application for safety-critical tasks, e.g., lane detection for autonomous driving.

*1) From lane detection to scene understanding*

Different from conventional objects, e.g., detecting dogs/cats and human faces, lane markings are frequently nicely structured. In most situations, lane lines appear as parallel polynomials evenly spaced on a relatively flat ground. This is largely true, except at lane merge/split, intersections, roundabouts, or on steep slopes. So if we can solve this "parallel polynomials" problem, we would have addressed the majority (> 90%) of the lane detection problem, which is the focus of this paper.

We are also inspired by the idea that we can deduce lanes from street scenes through low-level static or dynamic visual cues (e.g., lane markings, road curbs, and vehicles) and high-level functional cues (e.g., scene layout). Feature-based methods, such as ELAS [17], usually detect all possible scene cues in advance before lane detection/tracking task is executed. For Convolutional Neural Network (CNN)-based methods, such kind of scene information is hidden/implicit in the network architecture. If we can first understand the scene layout, separate the whole image into geometric areas and then focus on lane marking areas, the classification accuracy is expected improve.

Another important decision is what lane labels should the CNN generate. Many existing CNN-based methods generate pixel-wise lane marking flags, lane area masks or parameterized lane lines. Usually, manual inference for the occluded or missing parts are needed when labeling, thus uncertainty may be introduced in this process, see Fig. 1 (a-b). In addition, even more elaborate customized labels are required in some cases, e.g., VPGNet [15] outputs vanishing points in addition to lane

This research was supported by the UM-SFmotors automated car project and ARC project.
[1] P. Lu, S. Xu and H. Peng are with the Department of Mechanical Engineering and the Mcity, University of Michigan, Ann Arbor, MI 48109 USA (Email: pingpinl@umich.edu; xushao@umich.edu; hpeng@umich.edu)
[2] C. Cui and F. Wang is with the SF Motors Inc., Santa Clara, CA 95054 USA. (Email: cworkc56@gmail.com; fan.wang@driveseres.com)

attributes; LineNet [18] outputs six lane-related labels (mask, position, direction, confidence, distance, and type).

Instead of predicting the aforementioned task-oriented lane labels using CNNs, this paper attempts to solve the lane detection problem starting from scene understanding. Three benefits are highlighted here: 1) **Adaptability in complex scenarios**: compared with detecting lane objects directly, the holistic street scenes follow more stable layouts and are more robust to adverse factors such as lighting, occlusion and weather conditions. 2) **Reliability and reusability for perception**: we propose a hierarchical segmentation structure imitating human perception ability for reliable scene cues prediction, which can be reused in other perception tasks, e.g., drivable area detection in off-road driving. 3) **Compatibility with heterogeneous datasets**: the proposed structure works with different annotation categories, class definitions and labeling policies, see Fig. 1 (c). The above features made it possible to train on multiple datasets, e.g., Cityscape [19], Vistas [20], BDD [21], Camvid [22], KITTI [23], Apollo [24] and GTA5 [25]. These large-scale driving datasets could all be used to obtain better performance.

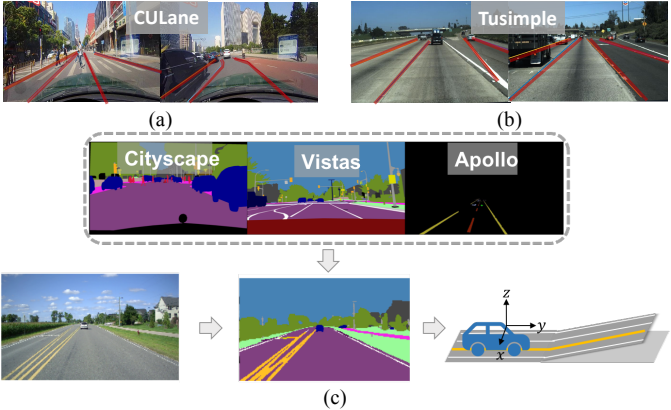

Fig 1. Comparison of labels among commonly used lane datasets and our training datasets. (a) (b) Examples with occluded/missing lane marks and lane labels; (c) Examples of our heterogeneous training datasets without labeling uncertainty, where the lane parameters are optimized based on scene labels without supervision.

*2) Lane parameter estimation*

For most CNN-based lane detection methods, the outputs are pixel-level lane instance masks in the image view [9, 13, 15]. However, the desired output for autonomous driving is control related parameters, i.e., vehicle lateral offset, heading angle and curvature [26]. To fill this gap, some post-processing procedures are required, e.g., inverse perspective mapping (IPM), and lane model fitting. Since the vehicle is moving and the road is not always flat, camera intrinsic and extrinsic parameters alone are not enough for calculating the transformation matrix between the camera and world coordinates. Better lane model or additional procedures, e.g., vanishing point estimation, are required [27]. It means more efforts are still needed after obtaining lane detection results in the image view. LaneNet-HNet [10] uses an extra CNN to estimate the transformation matrix from the image view to the birds' eye view (BEV) and conducts lane regression in BEV. LaneNet-LSTM [11] requires lane coordinates in BEV as input and identifies each lane line via LSTM. These estimation methods treat lane inference process as a highly nonlinear model to learn parameters in a supervised way.

Differently, we follow the assumption that lane markings are largely parallel polynomials, and if we separate the lane parameters into shared parts (heading angle and curvature) and unique parts (offsets and lane marking attributes), they can be estimated together efficiently. To cope with non-flat ground, a polynomial road model is adopted. Then the lane parameter estimation problem is simplified as an optimization problem based on the pixel-wise scene labels. The reasons are: 1) **to avoid step-by-step rule-based parameter estimation**, which increases problem complexity and causes error accumulation. Instead, we simultaneously estimate the slope angle and multi-lane parameters simultaneously; 2) **to leverage known attributes of lane lines.** We design an explicit cost function considering prior physics model/knowledge and optimize the lane parameters without supervision. We call this process "physics enhanced multi-lane inference".

The diagram of the proposed lane detection system is displayed in Fig. 2, including a hierarchical semantic segmentation module and a physics enhanced multi-lane inference module. The core idea is that CNNs are used for scene understanding as well as road/lane extraction, while a physical road/lane models are adopted for the lane inference. The explainability is improved, and the subsequent lane inference module uses a model and thus can be more accurate and reliable.

*B. Contribution*

The main contributions of this paper include:

1) **a novel lane detection system**, named as Scene Understanding Physics-Enhanced Real-time (SUPER). Its main difference from existing methods is that we solve the problem starting from scene understanding, and then estimate lane parameters through optimization with a physics-enhanced cost function, see Fig. 1.

2) **an improved hierarchical semantic segmentation structure** to capture lane related information focusing on the region of interest, i.e., where lane markings are likely to exist, and also enable training on multiple datasets.

3) **an optimization-based lane inference method** to directly estimate multi-lane parameters (i.e., lateral offsets, heading angle, and curvature) in real-time. Especially, a cost function considering road/lane models is used to formulate the parameter estimation problem.

The remaining contents are organized as follows: section II presents the proposed hierarchical semantic segmentation network, along with multi-level classifiers design and some training strategies; section III focuses on the details of lane parameter estimation, covering loss function design, slope compensation and optimization strategy; The experiments on open/private lane datasets, and real-world vehicle tests are conducted in section IV. Section V concludes this paper.

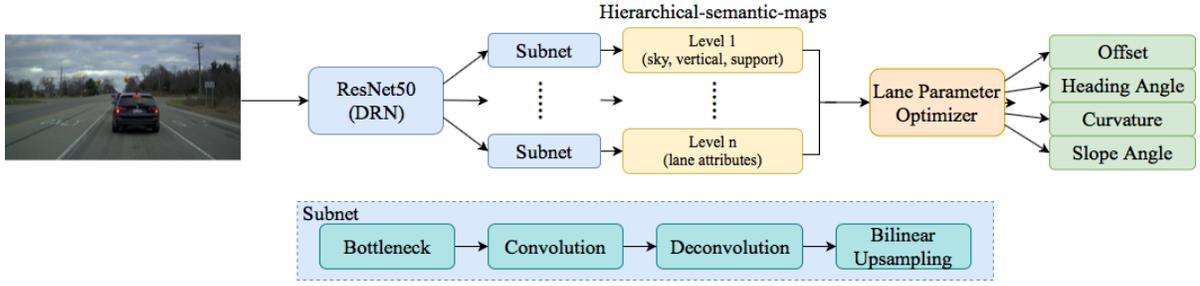

Fig. 2. Framework of the proposed method

## II. Hierarchical scene understanding

To understand the holistic scene for reliable lane detection, semantic segmentation is selected as the base network for the street scenes understanding. To capture the special semantic relationship between lanes/roads and other objects with multiple heterogeneous datasets, we modify the original hierarchical classification convolutional network [28] and utilize an improved semantic hierarchy by various lane attributes with a more practical architecture for lane detection. More details are explained in Sec. A and B. The training strategies are described in Sec. C.

### A. Hierarchical semantic segmentation

Semantic segmentation refers to the process of assigning a semantic label to each pixel of an image [29]. For learning based semantic segmentation, annotated training data is an essential part that affects model performance. In our hierarchical semantic segmentation model, both scene labels and detailed lane marking labels are required, while no current open dataset provides annotations covering all the required labels. To address this challenge, we propose the following concept to use multiple datasets to train our model.

Table 1. Training datasets used

|  | Cityscape[19] | Vistas[20] | Apollo[24] |
| --- | --- | --- | --- |
| Image size | 1024x2048 | varies | 2710x3384 |
| Training size | 2975 | 18000 | 7392 |
| Label | common scene | common scene | lane marking |
| Classes | 34 | 66 | 35 |

Three datasets are selected as the final training datasets, as shown in Table 1. Cityscape does provide semantic segmentation labels but not lane information. Vistas provides scene labels as well as some general categories of lane markings, such as 'bike lane', 'lane marking-general', etc. Apollo provides very detailed pixel-wise lane makings and lane attributes, including 6 dividing markings, 4 guiding markings, 2 stopping lines, 12 turning markings and so on. These three datasets together provided about 30,000 images with labels for our scene understanding based lane detection training and validation.

After gathering training datasets and labels, the next challenge is how to design the training process utilizing different datasets with varying labeling styles. Panagiotis et al. proposed a convolutional network with hierarchical classifiers for per-pixel semantic segmentation [28], which enables training on multiple heterogeneous datasets. Although many objects exist in the holistic scene, some of them are not related to lane detection and can be ignored, while others are highly related to lane features. Hence, we cannot use the structure of the algorithm proposed in [28] directly. To tackle the above issue, we optimize the hierarchy with a more practical architecture inspired by human perception ability. The relationships between different hierarchical levels are redefined, so that the proposed structure focuses on both general scene cues and lane classification. Our semantic hierarchical structure is explained below.

### B. Multi-level classifier design

Among available open datasets, the definition of object classes and labels are different, which needs to be clarified before they can be used. The main conflicts in the class definition are 'road' and 'lane marking'. For example, in Cityscape, 'road' refers to drivable ground, including lanes, direction arrows, streets, bicycle lanes, etc. In Vistas, lane markings are isolated from 'road'. To tackle 'road' conflict, a 'drivable' class is proposed which partially alleviate the issue, but other conflicts (e.g., types of vehicles) still exist. Besides, the designed network in [28] aims to predict many classes of objects, e.g., animals and birds, which will increase model complexity and training cost.

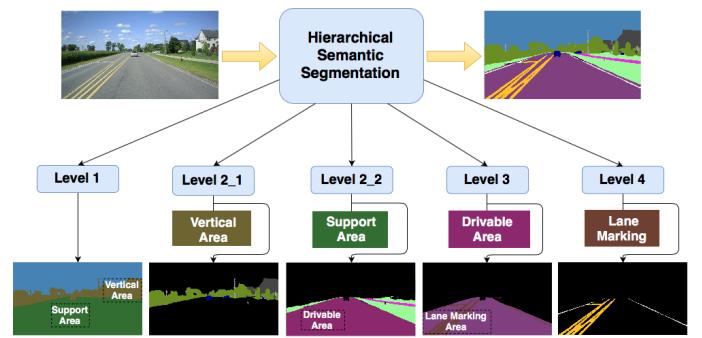

Fig. 3. The proposed hierarchical semantic structure. Support area includes sidewalk, ground, terrain, curb, and drivable areas; vertical area includes person, car, building, wall, bridge, tunnel, fence, vegetables and road utility; drivable area includes road, parking, crosswalk, lane marking, bike lane, service lane, catch basin, manhole and pothole; lane marking area includes dividing/guiding/turning markings, and stop lines.

To cope with labeling differences/conflicts and removing irrelevant labels, a new hierarchical structure is proposed. The idea is inspired by [30], which tries to label an image of a street scene into coarse geometric classes that are useful for tasks such as navigation, object recognition, and general scene understanding. Following the backbone network in Fig. 2, four-level classifiers are designed in our paper: in the first level, an image is divided into sky, vertical and support areas; in the second level, vertical/support areas are further divided into subclasses for scene description; in the third level, drivable areas are further classified; in the last level, detailed lane marking types are incorporated. Fig. 3 illustrates and explains the detailed multi-level structure.

*C. Training strategy*

Since only pixel-level classification is involved in this section, the loss function is relatively simple compared with [28]. Softmax cross-entropy loss is adopted for each classifier and the total loss is the weighted sum of different classifiers. In the inference procedure, each child classifier is governed by its parent according to its own decisions. Through this way, we obtain more reliable scene labels.

In the Apollo dataset, lane marking labels are more accurate than the ground and object labels, thus only lane segmentation label is used in the training stage. However, some ground area near the lane markings is easily misrecognized as lane markings. To handle this issue, we expand the lane marking to its vicinity, and label the whole expanded area as 'road'. Road areas occluded by objects such as vehicle are removed from the label. Fig. 4 exhibits an example of Apollo label before (left) and after (right) our expansion processing. Both lane markings and generated 'road' labels are used for training.

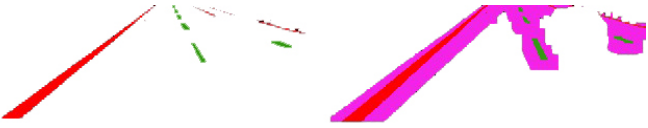

Fig. 4. Preprocessing of the Apollo lane dataset

### III. PHYSICS ENHANCED MULTI-LANE INFERENCE

With the semantic labels in Fig. 3, this section estimates lane parameters in the vehicle coordinate system. We assume the lane lines are largely parallel polynomials, and the lane parameters can be divided into two parts, shared/global parameters (i.e., heading angle and curvature) and unique/local parameters (i.e., offsets and lane marking attributes). In this section, the optimization of multi-lane parameters is described first. Then slope compensation (for up/down road grade) is proposed for better accuracy.

*A. Lane parameter optimization*

This section describes the lane parameter estimation as an optimization problem. The cost function is designed to reflect the physical properties of lane lines. The coordinates for the image view and BEV are illustrated in Fig. 5, along with definition of related symbols in the following.

The flat ground case is first considered here. Given intrinsic and extrinsic camera parameters, any road/lane points could be transformed from $(u^i, v^i)$ in the image view to $(x^i, y^i)$ in the BEV. Usually, a lane line can be represented by a polynomial function: $f(a_0^n, \boldsymbol{a}; y) = a_0^n + a_1 y + a_2 y^2 + \cdots + a_m y^m$, where $\boldsymbol{a} = \{a_1, a_2, \ldots, a_m\}$ contains the shared parameters; $a_0^n$ are the unique parameters – the offset of the $n^{th}$ lane line; $m$ is the polynomial order and chosen to be 2 in this paper. Two kinds of labels are involved: road labels $(x_r^i, y_r^i)$ and lane labels $(x_l^i, y_l^i)$.

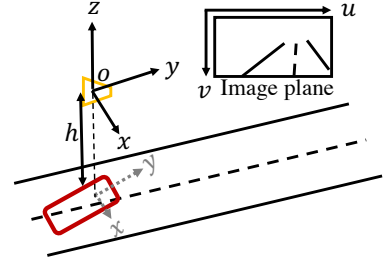

Fig. 5. Coordinates for the image view and BEV. Red rectangle stands for the vehicle, $o$ is the center of the camera, $h$ is the mounted height of camera above the ground.

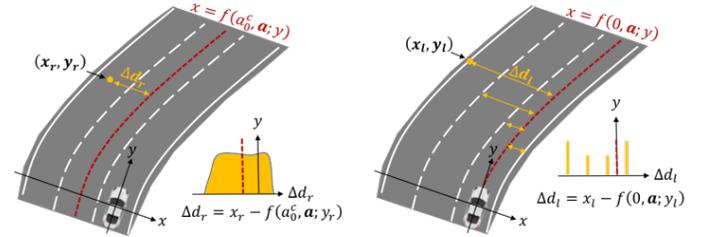

Fig. 6. Cost function for road fitting (left) and lane fitting (right)

The goal of the optimization problem is to estimate both $\boldsymbol{a}$ and $\{a_0^n\}$, where $n = 1, 2, \ldots, N$. $N$ is the total number of lane lines in a given image. To estimate the shared parameter $\boldsymbol{a}$, two separate cost functions considering the physics knowledge of road/lane are proposed and explained, see Fig. 6 for details.

**Road fitting**: assuming that (1) all lanes are in the road area; (2) the road area is represented by its central line; and (3) the lane lines share the same $\boldsymbol{a}$ as the central line of the road. We understand that assumption (3) is true for the majority of the cases, but exceptions exist (lane merge/split, intersection, etc.) Here we first aim to have a tool that helps us to solve 90% of the lane detection problem. To match the central line with the pixels of the road area, Eq. (1) is designed to indicate the distance between each point in the road area and the related central line. Minimizing $\mathcal{J}_1$ helps to find the best parameters, $a_0^c$ and $\boldsymbol{a}$.

$$\mathcal{J}_1 = -\sum_i exp\left\{-\left(\frac{|x_r^i - f(a_0^c, \boldsymbol{a}; y_r^i)|}{\sigma_d}\right)^2\right\} \quad (1)$$

where $a_0^c$ is the offset of the central line; $x_r^i$ and $y_r^i$ are the coordinates of each point in the road area; $\sigma_d$ is the standard deviation to penalize points far away from the central line, which can be set according to the road width. In this paper, $\sigma_d = 3$. This road-related cost function evaluates parameters $\boldsymbol{a}$ independent of $\{a_0^n\}$, $n \in [0, N]$.

**Lane fitting:** assuming that all parallel lane lines in the road areas share the same global geometric parameter $\boldsymbol{a}$. For a given point $(x_l^{i,n}, y_l^{i,n})$ on the $n^{th}$ lane lines, $\Delta d_l^{i,n} = x_l^{i,n} -$

$f(0, \boldsymbol{a}; y_l^{i,n})$ represents a value close to the offset of that lane. Eq. (2) then measures the distribution of the distance

$$\mathcal{J}_2 = Entropy\big(hist\{x_l^i - f(0, \boldsymbol{a}; y_l^i)\}\big) + \kappa \frac{N_l}{n_l} \quad (2)$$

where $x_l^i$ and $y_l^i$ are the $i^{th}$ coordinates of each sampled point in the lane marking areas; $N_l$ is the total number of sampled points in the lane marking category; $n_l$ is the number of lane marking points whose offsets fall into a pre-defined range, $\kappa$ is a constant and set to be 2 in this paper; $\kappa N_l/n_l$ is a penalty term for abnormal $\boldsymbol{a}$, $hist\{x_l^i - f(0, \boldsymbol{a}; y_l^i)\}$ stands for the distribution of offset $\Delta d_l^{i,n}$. Recall the parallelism assumption, $\Delta d_l^{i,n}$ on the same $n^{th}$ lane line should be the same or similar to $a_0^n$, thus a peak should occur around the correct $a_0^n$ value in the histogram. Considering all candidate points in the entire scene, several discrete peaks, standing for different offsets of all lane lines, are supposed to appear in the histogram. Ideally, the number of peaks is equal to lane number $N$. If we use map information, the correct value of $N$ can be known and used for more accurate detection.

Note that a disordered histogram indicates an inaccurate $\boldsymbol{a}$. To measure the level of error or the quality of $\boldsymbol{a}$, we introduce an index called $Entropy$, which calculates the entropy of $\{\Delta d_l^{i,n}\}$ through its corresponding histogram. A better $\boldsymbol{a}$ corresponds to a histogram that is closer to the ideal distribution with a lower entropy, while a higher entropy indicates inaccurate $\boldsymbol{a}$.

Eq. (1) and Eq. (2) utilizes the physics features of the road and lane to help with lane detection. With enough lane marking points, $\mathcal{J}_2$ of lane fitting contributes more than $\mathcal{J}_1$, whereas $\mathcal{J}_1$ of road fitting enables acceptable performance even without many lane points. To achieve a comprehensive evaluation of lane parameters, we combine them together to form the final loss function $\mathcal{J}$:

$$\arg\min_{a_0^c, \boldsymbol{a}} \mathcal{J} = \lambda \mathcal{J}_1 + \mathcal{J}_2 \quad (3)$$

where $\lambda$ is a weight and set to be 0.001 in this paper. Here the global parameter $\boldsymbol{a}$ is what we care about. The calculation of $\boldsymbol{a}$ turns into finding the best $\boldsymbol{a}$ and $a_0^c$ which minimizes $\mathcal{J}$.

Once the optimal $\boldsymbol{a}$ is obtained, the optimal distribution of $\{\Delta d_l^{i,n}\}$ is achieved as well as $\{a_0^n\}$, being the peaks of the best histogram

$$\boldsymbol{a_0} = peaks\big(hist\{x_l^i - f(0, \boldsymbol{a}; y_l^i)\}\big) \quad (4)$$

where $\boldsymbol{a_0} = \{a_0^0, a_0^1, \ldots, a_0^N\}$. The peak points were found using the *findpeaks*[1] function, which finds local maxima through setting minimum peak separation (equals the minimum road width in real world), and minimum peak prominence (equals the minimum number of points in each lane line). The first parameter is used to incorporate road width prior information and the second parameter is used to exclude interference of lanes in other directions. To better understand this process, a typical image along with the detailed lane offset estimation is shown in Fig. 7.

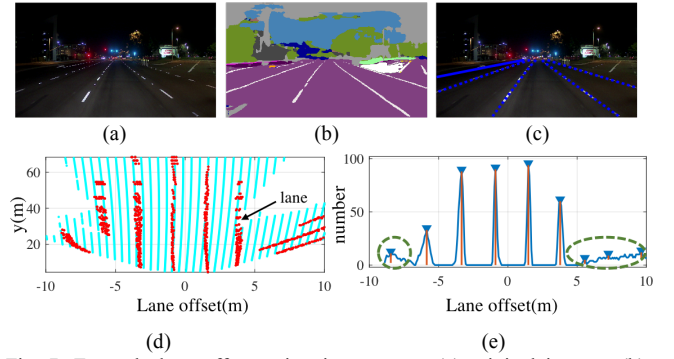

Fig. 7. Example lane offset estimation process. (a) original images; (b) semantic segmentation results; (c) lane detection results; (d) lane/road points after removing the global geometry via $x^i - f(0, \boldsymbol{a}; y^i)$, where the red/cyan points are for lane/road points, respectively; (e) lane offset estimation through peaks finding, where the blue line depicts the histogram corresponding to the x-axis value of the lane points in (d), the triangles indicate the peaks of the blue line after considering the minimum peak interval, red lines display the peak prominence, and the peaks in green ellipses are discarded after considering the minimum peak prominence.

### B. Slope compensation

In the real world, terrain undulations deteriorate the lane conversion from the image view to the BEV if only the default intrinsic and extrinsic camera parameters are used. On sloped roads, lane lines that are parallel in the real world become non-parallel in the BEV. To tackle this challenge, methods estimating vanishing point have been proposed [15, 27, 31]. The results are then used to update extrinsic camera parameters online. Differently, in this paper we propose a method to compensate for road slope in the optimization process directly.

Based on Inverse Perspective Mapping (IPM), a point $(u, v)$ from the image view can be transformed to $(x, y)$ in the BEV [32] by

$$\frac{1}{z_c}\begin{vmatrix}x\\y\\z\end{vmatrix} = T \begin{vmatrix}u\\v\\1\end{vmatrix} \quad (5)$$

where $T$ is the transformation matrix determined by extrinsic and intrinsic camera parameters, and $z_c$ is the distance along the z axis in the camera coordinate.

Since $z_c$ is unknown, the exact $(x, y, z)$ cannot be calculated from Eq. (5). Assuming the vehicle runs on a flat road, an estimated coordinate can be obtained, denoted as $(\tilde{x}, \tilde{y})$, and its relationship with $(x, y, z)$ follows

$$\begin{aligned}\tilde{x} &= -h\,x/z\\ \tilde{y} &= -h\,y/z\end{aligned} \quad (6)$$

where $h$ is the vertical height from the camera center to the flat ground.

If the road is not flat, we must estimate the z profile. Here we adopt a polynomial road model, assuming $z$ is a polynomial function related to $y$: $z = f_z(y)$. Then Eq. (6) becomes

$$\tilde{x} = -h\,x/f_z(y) \quad (7)$$

---
[1] https://www.mathworks.com/help/signal/ref/findpeaks.html

$$\tilde{y} = -h\,y/f_z(y)$$

If $f_z$ is known, Eq. (7) can then be rewritten as

$$\tilde{y} + h \cdot y/f_z(y) = 0 \tag{8}$$

Thus $y$ can be estimated from Eq. (8), denoted as $root(\tilde{y})$. Then, feed $root(\tilde{y})$ into Eq. (7) to obtain $x$

$$x = -\tilde{x} \cdot f_z\bigl(root(\tilde{y})\bigr)/h \tag{9}$$

Specially, assume $f_z$ is a linear function $f_z = by - h$. This simplified road model is suitable for most situations, except for frequently changing slopes. Then the correct $x$ and $y$ can be represented as

$$\begin{aligned} x &= -\tilde{x} \cdot f_z\left(\frac{\tilde{y}}{\tilde{y}b/h + 1}\right)/h \\ y &= \frac{\tilde{y}}{\tilde{y}b/h + 1} \end{aligned} \tag{10}$$

Note that $b$ is the only unknown parameter in Eq. (10). We treat it as an additional variable in the optimization problem, and optimize it together with the global lane parameters ***a***, and therefore the effect of the slope is estimated and mitigated.

### C. Optimization strategy

Since the derivative of the proposed loss function is not available, we use a Derivative-Free Optimization (DFO) algorithms [33], and more specifically, the Nelder-Mead simplex algorithm [34].

Usually, a proper initial guess helps to avoid trapped at the local minima and speed up the searching process. Sequential information may provide a good initial parameter, which means the optimal solution obtained from the previous images can be used as the initial value for the next image; curvature information of the road ahead the vehicle from map or navigation software is also beneficial. If none of them are available, other techniques can also be applied. Even when no prior information is available, a proper range can be selected for each parameter. Then follow Eq. (3) to calculate the loss curve of each parameter while setting other variables as 0.

## IV. EXPERIMENTAL VALIDATION

To evaluate the performance of the proposed lane detection algorithm, experiments on various datasets are performed. As mentioned before, our training datasets include Cityscape, Vistas and Apollo. While, the evaluation datasets include Tusimple, Caltech and X-3000.

Our algorithm was implemented using two platforms: Titian Xp for model inference, and Intel (R) Xeon (R) W-2155 CPU @ 3.30GHZ for lane parameter optimization. In the first module, the hierarchical model achieves an inference rate of 18 FPS. For the lane parameter optimization module, the average computation time is around 37 milliseconds. In other words, using today's computation platforms, we can achieve a combined computation rate of 11 Hz, capable of real-time vehicle implementation.

### A. Tusimple benchmark

Tusimple is a widely used open-source lane detection dataset, which consists of 3,626 training and 2,782 testing images. It mainly covers highway driving in good or fair weather conditions. Moreover, Tusimple provides video clips (20 frames per clip), but only the last frame of each clip is annotated.

The quantitative evaluation results are listed in Table 2, calculated by the official Tusimple benchmark evaluation script, compared with two selected benchmarks: SCNN [9] and LaneNet [10]. It is observed that joint estimation of slope and lane parameters achieves better detection accuracy (about 3% higher) than without, which confirms the necessity of slope compensation. Even though the accuracy is slightly worse than SCNN, the winner of the Tusimple 2017 competition, and the LaneNet, it is important to note that both SCNN and LaneNet were trained on the Tusimple data, whereas, the proposed method was not. In addition, the proposed method outputs lane parameters in the real world, and the results are then inversely mapped to the image view for the comparison; but SCNN and LaneNet focus on lanes in the image view without disturbance of road slope or mismatch of camera parameters. Besides, the accuracy is slightly improved from 95.90% to 96.01% after considering sequential information. This seems to imply that the driving data in Tusimple was very "normal" and typical default initial guess was close enough to the true solution. In this paper, temporal integration is simply performed through setting the optimization output of the previous frame as the initial guess of the current frame. In practical applications, multi-frame fusion could be considered, and advanced spatial or temporal filtering might be applied as well.

Table 2. Evaluation on the Tusimple test dataset, sc means using slope compensation for joint estimation, sq means incorporating sequential information for initial value setting

|  | SUPER | | SCNN | LaneNet |
|---|---|---|---|---|
|  | no sc | sc |  |  |
| no sq | 93.56% | 95.90% | **96.53%** | 96.4% |
| sq | 93.65% | 96.01% |  |  |

### B. Caltech benchmark

The Caltech dataset contains 1,224 labeled images with 4,172 marked lanes [7] from four video clips collected on different types of urban streets. For a fair comparison with other methods, we use the same evaluation metric – true positive rate (TPR) used in [8]. Table 3 shows the comparison results among several benchmark methods. Our method achieves the best performance in each sub-dataset. The average TPR is 98.6% which is over 10% improvement than the other two methods.

Table 3. Evaluation on Caltech dataset using True Positive Rate (TPR)

|  | 1 | 2 | 3 | 4 | Ave. |
|---|---|---|---|---|---|
| M. Aly[7] | 0.813 | 0.839 | 0.934 | 0.890 | 0.869 |
| ML-CRF[8] | 0.892 | 0.865 | 0.850 | 0.898 | 0.876 |
| SUPER[sc] | **0.991** | **0.980** | **0.982** | **0.992** | **0.986** |

## C. KITTI benchmark

The URABN KITTI-ROAD dataset [23] consists of 600 frames collected on five different days with relatively low traffic density. This paper focuses on the ego-lane detection subtask, which contains 100 testing images and 98 training images. Table 4 compared the results of our method in BEV with other methods. Since the ego-lane benchmark is not available in the testing images, the proposed method is evaluated on the training dataset using the official evaluation code, and other methods in Table 4 are evaluated on the testing dataset.

It is important to note that SPRAY [35], RBNet [36] and NVLaneNet[2] are all trained on the KITTI dataset. Even though KITTI's labeling style (ego-lane area) is quite different from that (lane line) used by Tusimple or Caltech, the proposed method without training on KITTI still achieves comparable performance.

Table 4. Evaluation on KITTI ego-lane dataset

|  | $F_{max}$ | Prec. | Rec. | FPR |
|---|---|---|---|---|
| BL[23] | 74.4 | 72.6 | 76.2 | 4.8 |
| SPRAY[35] | 83.4 | 84.8 | 82.1 | 2.6 |
| RBNet[36] | 90.5 | **94.9** | 86.6 | **0.82** |
| NVLaneNet* | **91.9** | 90.9 | **92.9** | 1.6 |
| SUPER[sc] | 86.7 | 86.0 | 87.4 | 2.2 |

* unpublished method

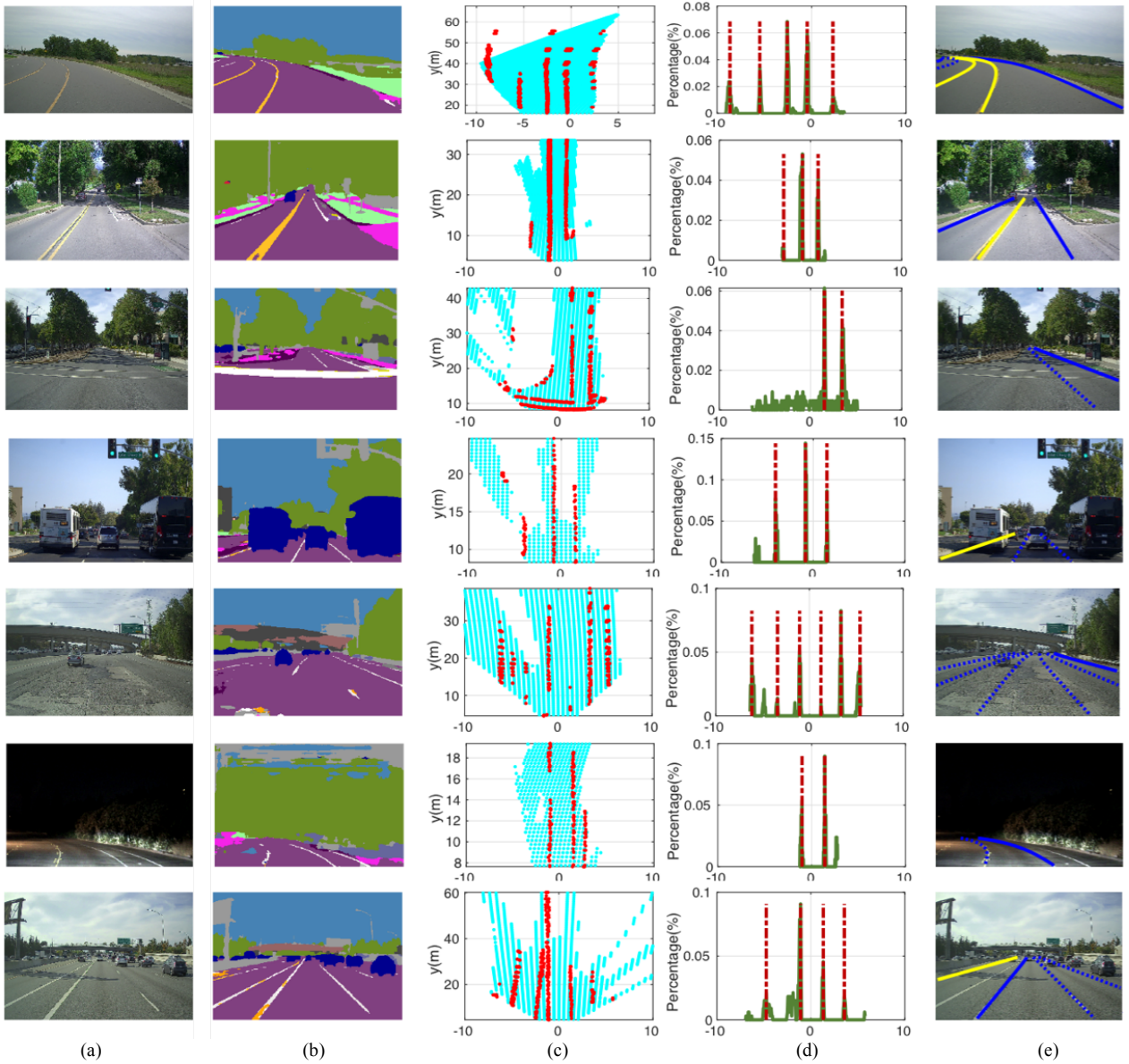

(a) (b) (c) (d) (e)

Fig. 8. Example results of the proposed method on the X-3000 dataset, (a) raw images, (b) semantic segmentation results, (c) offsets after slope compensation (red for lane points, cyan for road points), (d) histograms (green lines) of lane offsets and the detected peaks (red dashed lines), and (e) lane detection results, where blue/yellow stands for white/yellow lane line, line styles are also marked through solid/dashed lines.

[2] http://www.cvlibs.net/datasets/kitti/eval_road.php

*D. X-3000 benchmark*

X-3000 is our own dataset collected in urban, suburban and highway scenarios with different weather and time in both Michigan and California. X-3000 has three subsets: easy, moderate and hard, consisting of 1000 images each. The images are classified into these three challenging levels subjectively based on image quality and road/weather/lighting conditions. The easy subset covers straight and curve roads in good and fair weather; the moderate subset contains eroded or occluded lane markings, heavy shadow, uphill or downhill terrains and crossroad; the hard subset focuses on bad weather, bad image quality, complex intersection or lane merge/split. Our labeling style and assessment metric are similar to Tusimple. We use two evaluation modes: one focuses on ego-lane, and the other one evaluates 3-lanes (ego-lane and its left/right adjacent lanes). SCNN and LaneNet are as-is: trained on the Tusimple dataset, then tested on the X-3000 dataset. Sone typical results of our methods are displayed in Fig. 8, covering sharp curve, heavy occlusion, roundabout, intersection, split/merge, uphill/downhill, etc. Intermediate results, including semantic segmentation, offsets after global parameters removal and slope compensation, histogram of lane offsets, are also shown here.

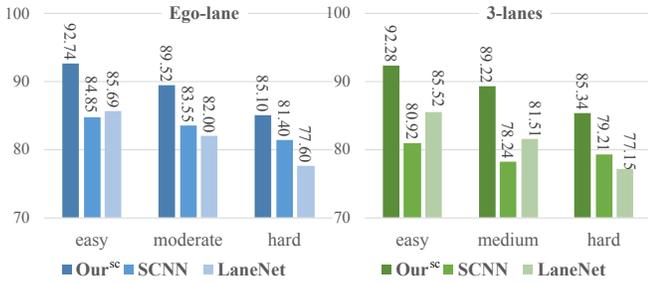

Fig. 9. Evaluation results using the X-3000 dataset

The accuracy of the two evaluation modes are summarized in Fig. 9. The average accuracy of our method is ~8 percentage higher than the others. Obviously, the proposed method outperforms the other two methods on these three subsets. Here, we also quantitatively compute the accuracy of lane color (white/yellow, 93.66%) and style (solid/dashed, 92.58%) evaluated with all three subsets. Note that all three methods were never trained on the X-3000 dataset. Therefore, the comparison here seems to indicate better robustness of our proposed method.

*E. Vehicle Testing*

To further validate the performance of our algorithm on a test vehicle in real-time, we select an open road in Michigan and drive our testing vehicle, a Lincoln MKZ, in a morning after rain. Fig. 10 (a) shows the test vehicle, (b) shows the test vehicle trajectory in the XYZ profile (The origin is marked by red rectangles), and (c) displays some typical camera images. This road is selected because it contains sharp curves, lane split/merge, intersections and frequently up-and-down terrains. Since we cannot access the original video of Mobileye, another camera (PointGrey) is used and mounted slightly to the right (20 cm away) of Mobileye to capture videos. When the test vehicle is running, the output of Mobileye (lane parameters only), vehicle position from Real-Time Kinematic (RTK) positioning system, raw video of PointGrey and outputs of our algorithm are recorded.

The vehicle trajectory captures from RTK does not exactly align with the lane center, which means the ground truths of offset ($a_0^n$) and heading angle ($a_1$) are not available. However, the curvature ($a_2$) information is available as it is not sensitive to instantaneous tracking errors. Therefore, here we focus more on $a_2$ for comparison. Fig. 11 (a) shows a general comparison of curvature profiles from the RTK trajectory (ground truth), Mobileye, and our algorithm. Fig. 11 (b–d) displays the curvature profiles under selected sections. The curvature profiles of Mobileye and our method are similar and close to the ground truth. Note that our algorithm only uses sequential information for initial value setting, without any spatial or temporal filtering. It seems our method produce results comparable to that of Mobileye in this test.

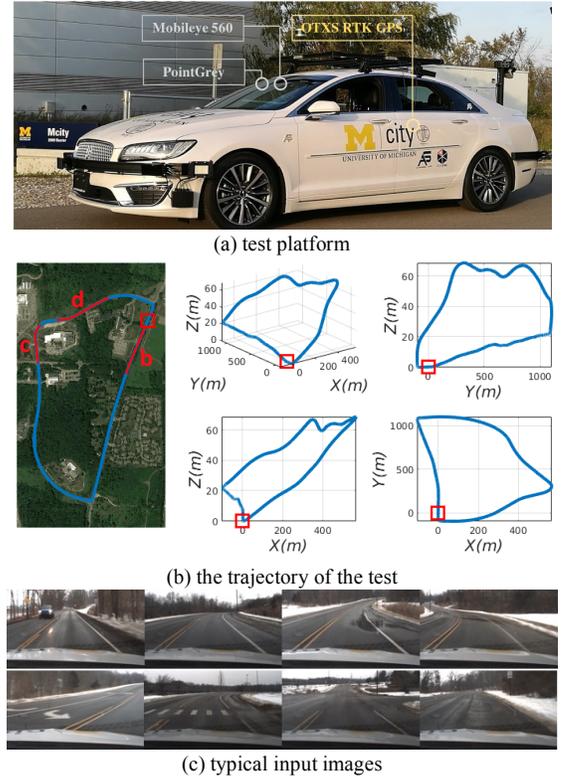

(a) test platform

(b) the trajectory of the test

(c) typical input images

Fig. 10. Test vehicle platform and test route

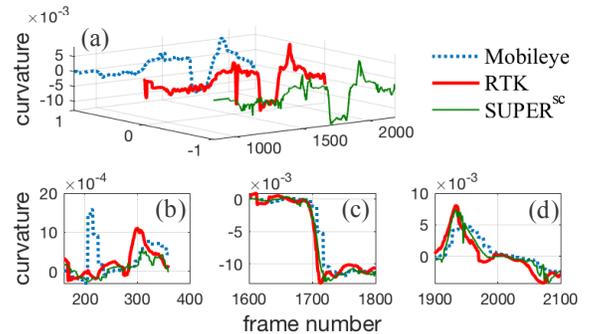

Fig. 11. Comparison of curvature profiles

## V. CONCLUSIONS

This paper proposed a novel lane detection algorithm with two unique ideas: 1) it predicts lane related labels from the holistic scene understanding; 2) it estimates multi-lane parameters and compensate for road slope simultaneously under an optimization framework. The advantages of both learning-based and physics-based techniques are leveraged.

The proposed algorithm is trained on heterogeneous datasets (Cityscape, Vistas, and Apollo) and then tested on four other datasets (Tusimple, Caltech, KITTI and X-3000). The proposed method was found to achieve similar or better performance, and is more robust. Comparison with Mobileye on open roads also indicates the performance of the proposed method seems to be fast enough for real-time implementation.

The lane inferring module presented in this paper follows the "parallel polynomials" hypothesis when optimizing lane parameters, which works well in most cases, but for accurate estimation of unparalleled lanes, such as lane merge and split conditions, additional operations/strategies, e.g., extra local correction or integration with map prior, are required, which is the focus of our current research.